\documentclass[11pt,a4paper]{article}
\usepackage[hyperref]{acl2021}
\usepackage{times}
\usepackage{latexsym}

\usepackage{microtype}

\aclfinalcopy 

\usepackage{microtype}

\usepackage[utf8]{inputenc} 
\usepackage[T1]{fontenc}    
\usepackage{hyperref}       
\usepackage{url}            
\usepackage{booktabs}       
\usepackage{amsfonts}       
\usepackage{nicefrac}       
\usepackage{microtype}      
\usepackage{lipsum}
\usepackage{multirow}
\usepackage{csquotes}  
\usepackage{verbatim}  
\usepackage{graphicx}

\title{\textsc{NukeLM}: Pre-Trained and Fine-Tuned Language Models for the Nuclear and Energy Domains}

\author{
    Lee Burke \and Karl Pazdernik \and Daniel Fortin \and Benjamin Wilson \and Rustam Goychayev
	\\
	Pacific Northwest National Laboratory \\ \AND
	John Mattingly \\
	North Carolina State University
	}

\date{}

\begin{document}
\maketitle


\begin{abstract}
  Natural language processing (NLP) tasks (text classification, named entity recognition, etc.) have seen revolutionary improvements over the last few years. This is due to language models such as \textsc{BERT} that achieve deep knowledge transfer by using a large pre-trained model, then fine-tuning the model on specific tasks. The \textsc{BERT} architecture has shown even better performance on domain-specific tasks when the model is pre-trained using domain-relevant texts. Inspired by these recent advancements, we have developed \textsc{NukeLM}, a nuclear-domain language model pre-trained on 1.5 million abstracts from the U.S. Department of Energy Office of Scientific and Technical Information (OSTI) database. This \textsc{NukeLM} model is then fine-tuned for the classification of research articles into either binary classes (related to the nuclear fuel cycle [NFC] or not) or multiple categories related to the subject of the article. We show that continued pre-training of a \textsc{BERT}-style architecture prior to fine-tuning yields greater performance on both article classification tasks. This information is critical for properly triaging manuscripts, a necessary task for better understanding citation networks that publish in the nuclear space, and for uncovering new areas of research in the nuclear (or nuclear-relevant) domains.
\end{abstract}


\section{Introduction}
\label{sec:introduction}

While natural language processing (NLP) has made significant strides in recent years, its application to the nuclear domain has remained rudimentary. In any domain, the ability to classify and triage information is critical when the data volume is large and growing. To enable the discovery of new connections between existing technologies or the potential use of a new technology in the nuclear domain, simple keyword searches are insufficient. To accelerate research in the nuclear domain, a  language model is needed—one that ``understands'' nuclear terminology, ``understands'' terminology in similar energy domains, and can automatically uncover latent similarities between materials, methodologies, and technologies.

In addition to accelerating nuclear science, this new methodology would be valuable to the International Atomic Energy Agency (IAEA) as part of their information collection and processing system. Quantifying the threat of a nation state's nuclear capability presents a particularly complex problem because the use, development, and transfer of nuclear technology is not itself an indication of nefarious intent. Technology itself has the added complexity of encompassing both physical items of trade, as well as social networks in academia and industry settings, where the “technology” is not a physical, tradeable good, but the knowledge and capabilities of individuals \cite{molas-gallart_which_1997}. Further, as international scientific collaborations become more prevalent, transfer of nuclear technology may become more prevalent, including inadvertent transfers. Readily available open-source information about such research collaborations, e.g., journal papers and technical reports, can offer indications of the use or transfer of such technology.

The amazing progress of state-of-the-art NLP methods has opened up new opportunities for nuclear domain researchers to leverage powerful language models. Models like \textsc{BERT} \cite{devlin_bert_2019} have shown significant improvement in NLP benchmarking metrics, such as the General Language Understanding Evaluation (GLUE) benchmark \cite{wang_glue_2019}. These benchmark metrics evaluate a language model's ability to perform a variety of tasks that resemble human ability to comprehend and be language literate. Though undoubtedly one element of \textsc{BERT}'s success is its large architecture of stacked Transformers \cite{vaswani_attention_2017}, another is the widespread use of transfer learning: pre-training on one task then fine-tuning on another. By pre-training on general-purpose corpora, a model has a strong foundation when approaching particular benchmark tasks.

There is also evidence that the performance of pre-trained language models on some tasks can be improved even further by domain-adaptive pre-training \cite{gururangan_dont_2020}---that is, starting with a model pre-trained on general-purpose corpora, then continuing the pre-training process on a corpus that is more representative of the domain of interest.

Given the recent success of large, Transformer-based neural network architectures and domain-adaptive pre-training, as well as the need for nuclear-``aware'' NLP models, we have developed \textsc{NukeLM}, a language model trained on nuclear-relevant research that performs best on nuclear-relevant downstream tasks.

\section{Related Work}
\label{sec:related-work}

A number of scientific and computational advances in recent years have led to significant improvements in the performance of computational models for natural language inference and understanding. Notable among these is the field of transfer learning, using pre-trained models for downstream tasks perhaps markedly different from their original tasks. Often, this takes the form of semi-supervised learning, where a model is trained on unlabeled data using a self-supervised task, then fine-tuned on a supervised task in the same domain.

Word embeddings (e.g., \textsc{word2vec} \cite{mikolov_distributed_2013}, \textsc{GloVe} \cite{pennington_glove_2014}, \textsc{fastText} \cite{mikolov2018advances}) learn a projection from the high-dimensional vocabulary space of a corpus of texts into a much smaller vector space using self-supervised training tasks like predicting nearby words. A key drawback of this approach is that each word is associated with a single vector, regardless of context.

A number of approaches have been proposed to learn contextualized word embeddings. For instance, \textsc{ELMo} \cite{peters_deep_2018} trains separate forward- and backward-oriented models for next-word prediction, then learns linear combinations of the deep representations for downstream tasks. In contrast, \textsc{BERT} \cite{devlin_bert_2019} learns to encode context from both left and right at once using a very large architecture of stacked Transformers \cite{vaswani_attention_2017}, pre-training with both a word prediction task (masked language modeling, MLM) and a task to predict whether a given sample follows another in the original text, relative to being chosen randomly from the corpus (next-sentence prediction, NSP).

\textsc{RoBERTa} \cite{liu_roberta_2019} leverages the same Transformer-based architecture as \textsc{BERT}, but shows improvements on downstream tasks with some changes to its pre-training strategy: it removes the NSP objective, pre-training only with MLM; it allow samples to cross document boundaries in pre-training, ensuring all pre-training samples are as long as possible; it determines which tokens to predict in each batch rather than deciding offline, before training; it uses much larger batch sizes; it uses byte-level tokenization instead of character-level; and finally, it considers much more pre-training data, including those from the Common Crawl corpora.

\textsc{SciBERT} \cite{beltagy_scibert_2019} clones \textsc{BERT}'s stacked Transformer architecture and pre-training methodology but replaces the \textsc{BERT} training corpus with a large, multi-domain corpus of scientific publications. This results in better performance on scientific domain tasks because of the better match between the domains of pre-training and fine-tuning tasks.

In contrast to training a domain-specific model from scratch like \textsc{SciBERT}, Gururangan et al. \cite{gururangan_dont_2020} demonstrate that continued pre-training of a general-purpose language model on in-domain text (called domain-adaptive pre-training, DAPT) can lead to improved performance on downstream tasks, but that continued pre-training on out-of-domain text can worsen performance. They explore several ways to bootstrap a targeted continued-pre-training corpus and explore the tradeoff between performance and computational expense.

Similarly, several domain-specific models have been proposed that continue pre-training from a \textsc{BERT} checkpoint. \textsc{BioBERT} \cite{lee_biobert_2019} continues pre-training on biomedical corpora. \textsc{NukeBERT} \cite{jain_nukebert_2020} continues pre-training on a nuclear-domain corpus, with the addition of newly initialized vocabulary entries specific to the nuclear domain. However, in contrast to \textsc{NukeLM}, the pre-training corpus for the \textsc{NukeBERT} model was generated from a relatively small corpus consisting of about 7000 internal reports from the Indira Gandi Center for Atomic Research, largely focused on fast breeder reactors; the \textsc{NukeBERT} language model is somewhat narrowly focused on nuclear reactor research for power generation rather than defining topics broadly associated with the nuclear fuel cycle. Furthermore, it is not clear if the \textsc{NukeBERT} language model is publicly available, and the associated dataset is not available under a standard open-source license.

\section{Data}
\label{sec:data}

We consider scientific abstracts from the U.~S.~Department of Energy Office (DOE) Scientific and Technical Information (OSTI) database \cite{OSTI} obtained in November 2018, amounting to nearly two million abstracts from over 70 years of research results from DOE and its predecessor agencies. 

For fine-tuning, we consider only abstracts labeled with a subject category. The possible categories are formalized by OSTI, and all products submitted to OSTI are encouraged to provide at least one, listing the primary category first. If more than one category is specified, we consider only the first.  In addition to the multi-class labels induced by the OSTI subject categories, we formulate binary labels by identifying OSTI subject categories that correspond to the top level of the IAEA Physical Model \cite{liu_development_2001}, which describes acquisition pathways. The topics described in the IAEA Physical Model include ore mining and milling, pre-conversion, uranium enrichment, post-conversion, fuel fabrication, nuclear reactors, heavy water production, and reprocessing of irradiated fuels. Using this criterion, the following OSTI topic categories are considered related to the nuclear fuel cycle for the binary classifier: nuclear fuels, isotope and radiation sources, nuclear fuel cycle and fuel materials, management of radioactive and nonradioactive wastes from nuclear facilities, specific nuclear reactors and associated plants, general studies of nuclear reactors, radiation chemistry, instruments related to nuclear science and technology, and nuclear physics and radiation physics. The list of all OSTI categories and their binary categorization designation is provided in Appendix A.

\section{Experimental Setup}
\label{sec:exp-setup}

We begin with pre-trained checkpoints implemented in HuggingFace's \texttt{transformers} framework \cite{Wolf2019HuggingFacesTS}, available from the HuggingFace model database with the following slugs: \texttt{roberta-base} and \texttt{roberta-large} are base and large versions of the \textsc{RoBERTa} model, respectively, and \texttt{allenai/scibert\_scivocab\_uncased} is the recommended uncased version (i.e., inputs are converted to lower case) of \textsc{SciBERT}.

Following \cite{gururangan_dont_2020}, we perform domain-adaptive pre-training. We continue pre-training all three models, \textsc{SciBERT}, \textsc{RoBERTa Base}, and \textsc{RoBERTa Large}, on 80\% of the OSTI abstracts, and for the remainder of this manuscript, we use the naming convention \textsc{NukeLM} to define the latter. The remaining 20\% of documents are held out from the pre-training process and split evenly into two data sets (~200 K each) to be used for fine-tuning and testing the classification models.  When forming each batch, 512-token segments are taken irrespective of document boundaries, and 15\% of the tokens are masked for prediction. We train for 13 K steps with a batch size of 256, for a total of 3.3 M samples consisting of 1.7 B tokens (similar in size to the corpora in \cite{gururangan_dont_2020}). Other hyperparameters follow \cite{gururangan_dont_2020}.

We perform some exploratory analysis of the impact of domain-adaptive pre-training on OSTI abstracts, including performance metrics and an example of masked word modeling.

For fine-tuning, we begin with the six models described above: \textsc{RoBERTa Base} and \textsc{Large} and \textsc{SciBERT}, both with and without OSTI domain-adaptive pre-training. We then follow \cite{gururangan_dont_2020} by passing the final layer \texttt{[CLS]} token representation to a task-specific fully connected layer for prediction (see the \texttt{transformers} documentation for details). A validation set is held out, consisting of 10\% of the overall fine-tuning set. 

We consider two tasks: multi-class prediction over the original OSTI subject categories, and binary prediction over the relevance of an abstract's subject category to one of the steps of the nuclear fuel cycle. The fine-tuning data set consisted of 198,564 documents, of which 23,268 are related to the nuclear fuel cycle according to our definition.

A small hyperparameter search is performed on the binary task (details in Appendix B), selecting a learning rate of $10^{-5}$ and a batch size of 64. We train for five epochs (14.7 K steps), evaluating at 20 checkpoints (about every 750 steps) and saving the best model according to loss on the validation set. Other hyperparameters follow \cite{gururangan_dont_2020}.


\section{Results of the Language Modeling Task}
\subsection{Metrics}
The MLM task is evaluated based on the categorical cross-entropy between the one-hot true distribution over a model's vocabulary and a model's predicted distribution. This MLM loss is shown before and after domain-adaptive pre-training for each of the three baseline models in Table \ref{tab:exp_1}.

Continued pre-training improves the performance of \textsc{RoBERTa Base} more than that of \textsc{SciBERT}, to the point where it performs better than the much larger \textsc{RoBERTa} without continued pre-training. The \textsc{RoBERTa} pre-training strategies may have yielded an easier-to-train model than the \textsc{SciBERT} methodologies, but this may be due solely to the larger vocabulary size, 50 K tokens for \textsc{RoBERTa} vs. 30 K for \textsc{SciBERT}. Regardless, \textsc{NukeLM} shows improvement over \textsc{RoBERTa Large}, and remains the most accurate of the models.

\begin{table}[htbp]
  \caption{Masked language modeling loss, based on categorical cross-entropy between true and predicted probability distributions, on the evaluation sub-set of the OSTI pre-training data. Lower is better. The symbol ``+ OSTI'' denotes continued pre-training on OSTI abstracts. The best performing model is in \textbf{bold}.}
  \small
  \centering
    \begin{tabular}{lll}
        \toprule
        Model            & MLM Loss \\
        \midrule
        \textsc{RoBERTa Base} & 1.39     \\
        \midrule
        \textsc{RoBERTa Base} + OSTI & 1.11     \\
        \midrule
        \textsc{RoBERTa Large} & 1.13     \\
        \midrule
        \textbf{\textsc{NukeLM}} & \textbf{0.95}     \\
        \midrule
        \textsc{SciBERT} & 1.34     \\
        \midrule
        \textsc{SciBERT} + OSTI & 1.18     \\
        \bottomrule
    \end{tabular}
  \label{tab:exp_1}
\end{table}

\subsection{MLM Example}
We present an example of masked language modeling to illustrate the task and performance improvement after domain-adaptive pre-training. The \textbf{bolded} word is masked, and the models are asked to predict what word should fill in the blank.

\begin{displayquote}
    The use of heavy \textbf{water} as the moderator is the key to the PHWR system, enabling the use of natural uranium as the fuel (in the form of ceramic UO2), which means that it can be operated without expensive uranium enrichment facilities. \cite{noauthor_pressurized_2020}
\end{displayquote}

Table \ref{tab:mlm_2} summarizes the top five predicted tokens and their associated likelihood score from each of the six models after domain-adaptive pre-training (if any) but before fine-tuning. Before continued pre-training, all three models include the correct answer in their top five predictions, but \textsc{RoBERTa Base} and \textsc{SciBERT} predict the more common but incorrect phrase ``heavy metal,'' albeit with low confidence; only \textsc{RoBERTa Large} predicts the correct answer. After continued pre-training, all three models succeed in predicting the correct answer with high confidence.

\begin{table}[htpb]
    \caption{An example of masked language modeling. Column 2 contains the top five tokens considered most likely (the true token, ``water'', is in \textbf{bold}), and column 3 contains the associated likelihood scores (the highest confidence for the true token is also in \textbf{bold}). The character ``\#'' indicates the token is a sub-word, i.e., a prediction of ``heavywater'' rather than ``heavy water''. The symbol ``+ OSTI'' denotes continued pre-training on OSTI abstracts.}
    \small
    \centering
    \begin{tabular}{llll}
        \toprule
        Model            & Top-5 Preds. & Score \\
        \midrule
                         & metal        & 0.252 \\
        \textsc{RoBERTa} & metals       & 0.149 \\
        \textsc{Base}    & uranium      & 0.145 \\
                         & \textbf{water}        & 0.130 \\
                         & iron         & 0.026 \\
        \midrule
                         & \textbf{water}        & 0.955 \\
        \textsc{RoBERTa} & metal        & 0.008 \\
        \textsc{Base}    & elements     & 0.008 \\
        + OSTI           & metals       & 0.008 \\
                         & oil          & 0.003 \\
        \midrule
                         & \textbf{water}        & 0.951 \\
        \textsc{RoBERTa} & metal        & 0.013 \\
        \textsc{Large}   & metals       & 0.011 \\
                         & fuel         & 0.004 \\
                         & carbon       & 0.002 \\
        \midrule
                         & \textbf{water}        & \textbf{0.996} \\
        \textsc{NukeLM} & metals       & 0.001 \\
           & oil          & 0.001 \\
                   & \#water      & <0.001 \\
                         & metal        & <0.001 \\
        \midrule
                         & metal        & 0.225 \\
        \textsc{SciBERT} & metals       & 0.117 \\
                         & \textbf{water}        & 0.068 \\
                         & iron         & 0.052 \\
                         & argon        & 0.042 \\
        \midrule
                         & \textbf{water}        & 0.929 \\
        \textsc{SciBERT} & metal        & 0.024 \\
        + OSTI           & metals       & 0.011 \\
                         & iron         & 0.003 \\
                         & oil          & 0.003 \\
        \bottomrule
    \end{tabular}
    \label{tab:mlm_2}
\end{table}

\section{Results of Downstream Tasks}

\subsection{Multi-Class Classification}
The results of fine-tuning of the multi-class classification task are presented in Table \ref{tab:fine_tune_multi}. \textsc{SciBERT}'s advantage over \textsc{RoBERTa Base} persists after domain-adaptive pre-training, perhaps because its scientific-domain pre-training corpora are more closely related to the OSTI task than are \textsc{RoBERTa}'s. However, neither overcomes \textsc{RoBERTa Large} even without the added advantage of continued pre-training, likely because the latter contains an order of magnitude more trainable parameters.

\begin{table}[htpb]
    \caption{Results of fine-tuning on the multi-class classification task. Precision, Recall, and F1 scores are an average of all classes, weighted by class size. The best performing model by each metric is presented in \textbf{bold}.}
    \small
    \centering
    \begin{tabular}{llllll}
        \toprule
        Model            & Accuracy & Precision & Recall & F1 score \\
        \midrule
        \textsc{RoBERTa} & 0.6745   & 0.6564 & 0.6745 & 0.6603 \\
        \textsc{Base}    &          &        &        &        \\ 
        \midrule
        \textsc{RoBERTa} & 0.6972   & 0.6884 & 0.6972 & 0.6863 \\
        \textsc{Base}    &          &        &        &        \\ 
        + OSTI           &          &        &        &        \\
        \midrule
        \textsc{RoBERTa} & 0.7056   & 0.7008 & 0.7056 & 0.7013 \\
        \textsc{Large}   &          &        &        &        \\ 
        \midrule
        \textsc{NukeLM} & \textbf{0.7201}   & \textbf{0.7164} & \textbf{0.7201} & \textbf{0.7168} \\
           &          &        &        &        \\ 
        \midrule
        \textsc{SciBERT} & 0.6972   & 0.6866 & 0.6972 & 0.6883 \\
                         &          &        &        &        \\
        \midrule
        \textsc{SciBERT} & 0.7047   & 0.6981 & 0.7047 & 0.6973 \\
        + OSTI           &          &        &        &        \\
        \bottomrule
    \end{tabular}
    \label{tab:fine_tune_multi}
\end{table}

\subsection{Binary Classification}
\label{sec:binary-class}
The results of fine-tuning on the binary classification task are presented in Table \ref{tab:fine_tune_binary}. Without domain-adaptive pre-training, \textsc{SciBERT} performs even better than \textsc{RoBERTa Large}, possibly because of its more closely related pre-training corpora. However, unlike in the multi-class task, both \textsc{SciBERT} and \textsc{RoBERTa Base} see degraded recall, outweighed by a moderate increase in precision only due to class imbalance. Only \textsc{NukeLM} sees improvement across all measured metrics, likely due again to its large size. It is worth noting that the much smaller \textsc{RoBERTa Base} is able to achieve performance comparable to the unwieldy \textsc{RoBERTa Large} via continued pre-training, which may be useful in resource-constrained applications.

\begin{table}[htbp]
    \caption{Results of fine-tuning on the binary classification task. Precision, Recall, and F1 scores consider NFC-related to be the positive class. The best performing model by each metric is presented in \textbf{bold}.}
    \small
    \centering
    \begin{tabular}{llllll}
        \toprule
        Model            & Accuracy & Precision & Recall & F1 score \\
        \midrule
        \textsc{RoBERTa} & 0.9506   & 0.7938 & 0.7816 & 0.7876 \\
        \textsc{Base}    &          &        &        &        \\ 
        \midrule
        \textsc{RoBERTa} & 0.9544   & 0.8237 & 0.7773 & 0.7998 \\
        \textsc{Base}    &          &        &        &        \\ 
        + OSTI           &          &        &        &        \\
        \midrule
        \textsc{RoBERTa} & 0.9506   & 0.7995 & 0.7722 & 0.7856 \\
        \textsc{Large}   &          &        &        &        \\ 
        \midrule
        \textsc{NukeLM} & \textbf{0.9573}   & 0.8270 & \textbf{0.8038} & \textbf{0.8152} \\
           &          &        &        &        \\ 
        \midrule
        \textsc{SciBERT} & 0.9548   & 0.8061 & 0.7910 & 0.7984 \\
                         &          &        &        &        \\
        \midrule
        \textsc{SciBERT} & 0.9532   & \textbf{0.8285} & 0.7747 & 0.8007 \\
        + OSTI           &          &        &        &        \\
        \bottomrule
    \end{tabular}
    \label{tab:fine_tune_binary}
\end{table}

\subsection{Performance under Different Training Set Sizes}
One reported advantage of domain-adapted languaged models is the ability to fine-tune on smaller numbers of labeled examples. We test this ability with the binary classification task described above.

\begin{figure}[htbp]
    \centering
    \includegraphics[width=0.95\linewidth]{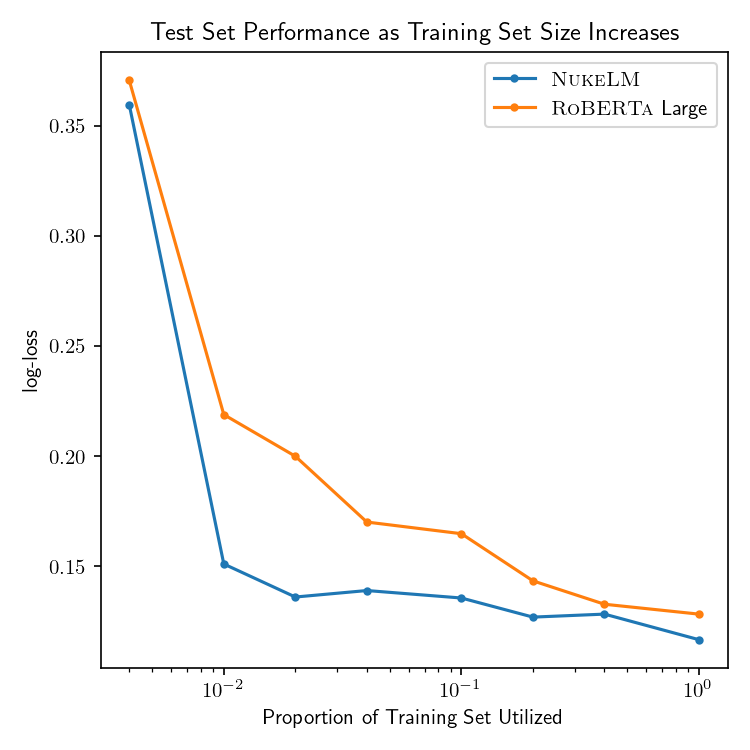}
    \caption{Binary classification performance, measured by log-loss on a hold-out evaluation set, as the training set size is increased, for \textsc{RoBERTa Large} (orange) and \textsc{NukeLM} (blue).}
    \label{fig:scaling}
\end{figure}

We randomly select increasingly large proportions of the binary classification fine-tuning set, ignoring the rest, so that each larger subset contains the earlier, smaller subsets. We train the off-the-shelf \textsc{RoBERTa Large} and \textsc{NukeLM} with the same experimental set-up as in Section \ref{sec:binary-class} and track the log-loss computed on the hold-out evaluation set. Figure \ref{fig:scaling} summarizes the results.

The domain-adapted model achieves significantly better performance with smaller amounts of data, however this advantage shrinks as the training set size increases. This could be the result of continued pre-training priming the model for performance in this domain. However, this pattern is not entirely consistent which is likely an effect of the random selection of documents.

Interestingly, and contrary to prior expectation, the disparity between models actually decreases at the lowest sample size tested (0.4$\%$ of the full corpus). While \textsc{NukeLM} maintains its superiority, with so few examples used for fine-tuning, neither model performs well. We suspect that at a sample size this low (754 versus 1000 documents used for training and test, respectively), both models may be overfitting to the limited trained data, which is why the dominance of \textsc{NukeLM} is muted.

\subsection{Qualitative Assessment}
Beyond model performance on the MLM task and document classification, an important question regarding these trained language models is whether or not any reasonable interpretation can be gleaned from the generally unintelligible vectors representations of the input text. While there is not a clear consensus on how useful these embeddings can be in providing explanations, with arguments from both sides \cite{jain_attention_2019,wiegreffe_attention_2019}, there is undoubtedly some information contained within these transformer-based language models because their predictive ability is state-of-the-art. So, while a direct interpretation of an embedding produced by \textsc{NukeLM} may be questionable, the transformation of this high-dimensional space that results from pre-training should provide some explanation as to why prediction was improved.

\begin{figure*}[htbp]
    \centering
    \includegraphics[width=0.95\linewidth]{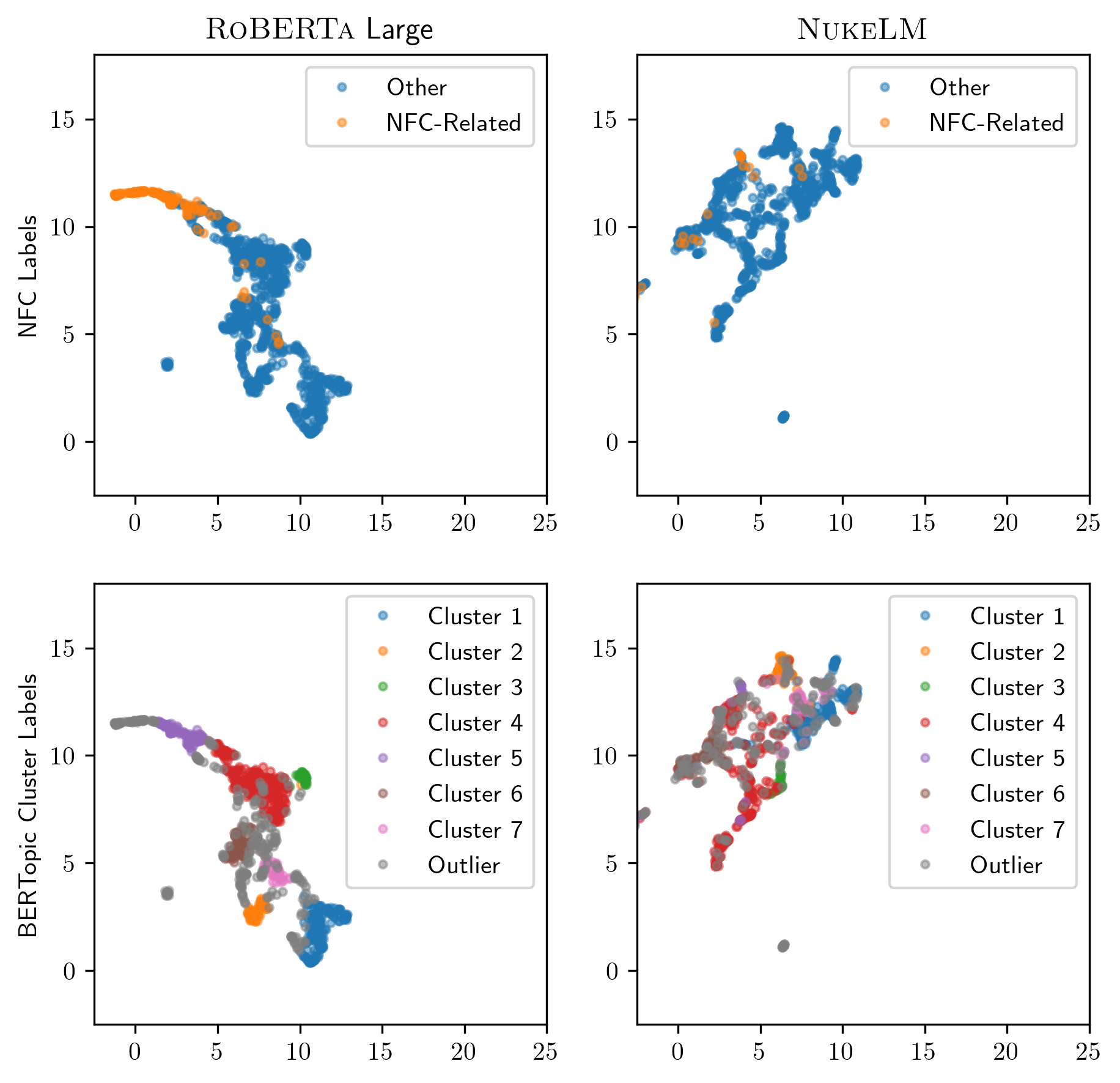}
    \caption{Visualization of UMAP-transformed output embeddings from \textsc{RoBERTa Large} for 1000 randomly sampled documents from the validation set after fine-tuning on the binary classification task, both without (left) and with (right) domain-adaptive pre-training on OSTI abstracts, colored by the true binary labels (top) and BERTopic clusters (bottom). Each point in these plots is a low-dimensional representation of the embedding for a document's abstract.}
    \label{fig:umaps}
\end{figure*}

As a first step toward interpreting the impact of domain-adaptive pre-training, we visualize output embeddings from the most accurate model, \textsc{RoBERTa Large}, both with and without continued pre-training on OSTI abstracts, and after fine-tuning on the binary classification task. We use UMAP \cite{mcinnes2018umap-software} with all default parameters to project the output corresponding to the special token \texttt{[CLS]} down to two dimensions, training separate UMAPs for each model. Figure \ref{fig:umaps} (top row) depicts the result of this process performed on a 1000-sample random subset of the binary classification task validation set.

In both models, the positive class is generally clustered together; indeed, both models are able to learn relatively accurate decision boundaries. However, in the version without domain-adaptive pre-training, the cluster looks like a single manifold, eventually connecting to the mass of negative samples like an isthmus. In contrast, continued pre-training appears to encourage the model to form more complicated structures, with an isolated cluster of mostly positive samples in addition to a similar but much smaller isthmus connected to a large mass of negative samples.

To explore these differences further, we apply BERTopic \cite{grootendorst2020bertopic}, a clustering and topic modeling approach for understanding the output embeddings of a transformer model. BERTopic also uses a UMAP projection for dimension reduction, in this case to 100 dimensions, but then uses HDBSCAN \cite{campello_density-based_2013} to cluster documents and a class-based TF-IDF (cb-TF-IDF) score for topic modeling. Here, all documents within the same cluster are concatenated into a single document and then the usual TF-IDF score \cite{teller2000speech} is computed as follows:

\begin{equation}
    cb-TF-IDF_i = \frac{t_i}{w_i} \times \log\frac{m}{\sum_{j=1}^n t_j},
\end{equation}

\noindent where $t_i$ is the frequency of each word in class $i$, $w_i$ is the the total number of words in class $i$, $m$ is the number of documents, and $n$ is the number of classes.

We visualize the BERTopic clusters found in the \textsc{RoBERTa Large} binary classification models in Figure \ref{fig:umaps} (bottom row). The three words most representative of each cluster, as determined by the cb-TF-IDF model, are listed in Table \ref{tab:bertopic-topics}. Without continued pre-training, we see seven clusters on a variety of topics, from cosmology to biology, with the NFC-related samples mostly relegated to a single nuclear cluster or left as outliers. In contrast, with continued pre-training, non-NFC topics are grouped into a single cluster and nuclear topics are given two clusters, one focused on nuclear facilities and another on nuclear physics. This provides evidence that continued pre-training taught the model additional knowledge of the nuclear domain, allowing it to characterize different subsets of positive examples, and recognize the irrelevance of other distinctions to the fine-tuning task.

\begin{table}[htbp]
    \caption{Top three representative words for each BERTopic cluster of output embeddings from \textsc{RoBERTa Large} for 1000 randomly sampled documents from the validation set after fine-tuning on the binary classification task, both without and with domain-adaptive pre-training on OSTI abstracts. Column two, the cluster number, corresponds with the legends in Figure \ref{fig:umaps} (bottom row).}
    \small
    \centering
    \begin{tabular}{llllll}
        \toprule
        Model                          & No. & Top-3 Words \\
        \midrule
                         & 1           & galaxies, galaxy, dust \\ 
                         & 2           & coal, gas, energy \\ 
        \textsc{RoBERTa} & 3           & cells, protein, kinase \\ 
        \textsc{Large}   & 4           & beam, laser, plasma \\ 
                         & 5           & waste, nuclear, radiation \\ 
                         & 6           & alloy, temperature, conductivity \\ 
                         & 7           & water, alkanolamine, lake \\ 
        \midrule
         & 1           & nuclear, reactor, facility \\ 
        \textsc{NukeLM}   & 2           & energy, system, model \\ 
                  & 3           & nuclei, neutron, energies \\ 

        \bottomrule
    \end{tabular}
    \label{tab:bertopic-topics}
\end{table}

\section{Conclusion and Future Work}
\label{sec:conclusion}

In this work, we leveraged abstracts from the OSTI database to train state-of-the-art language models for nuclear-domain-specific classification tasks and as a general-purpose language model in the nuclear domain. We explored a number of base models for transfer learning and applied domain-adaptive pre-training to improve performance on the down-stream tasks. To the best performing model in this process, \textsc{RoBERTa Large + OSTI}, we apply the name \textsc{NukeLM}.

We consider the \textsc{NukeLM} language model to be a general purpose resource for supporting development of NLP models in the nuclear domain. The \textsc{NukeLM} model can be leveraged for task training on relatively small labeled data sets, making it feasible to manually label training for targeted objectives and easily fine-tune the \textsc{NukeLM} model for various tasks.

Future versions of \textsc{NukeLM} will improve the model training pipeline, including considering full article text and data sets other than OSTI, expanding the model vocabulary to better capture a nuclear domain vocabulary without losing \textsc{RoBERTa}'s more robust pre-training, and exploring multilingual capabilities via models like \textsc{XLM-RoBERTa} \cite{conneau_unsupervised_2020}.

We introduced a binary categorization of the OSTI subject categories aimed at identifying documents related to the nuclear fuel cycle and fine-tuned the \textsc{NukeLM} model on this task. This fine-tuned classification model can be immediately useful as triage tool or to support NLP workflows in nuclear science or nuclear nonproliferation. 

The \textsc{NukeLM} binary classification model demonstrated superior performance for the classification task, though performance gains were minor. Because of computational constraints, multiple runs of the training process were not made to establish the statistical significance of the classification metrics, but the large set of training data and consistent trends across model types and tasks make it unlikely that the rank order of these models would change with resampling and retraining. Furthermore, we demonstrate that the performance gain may be even higher with smaller-scale fine-tuning sets.

Moreover, the whole story does not lie within the F1-score, because our qualitative visual assessment of the \textsc{NukeLM} binary classification embeddings reveal intriguing structural differences. The \textsc{NukeLM} embeddings appear to have more distinct clusters and increased separation among clusters, particularly among NFC-related documents. By applying BERTopic to these embeddings, we confirmed that these clusters correspond to identifiable topics. Potential future work would be needed to quantify these structural changes and assess differences among various models, as an in-road toward explaining how the models reach their conclusions.

\section*{Acknowledgments}
The authors thank Aaron Luttman and Matthew Oster for their helpful feedback; and Gideon Juve, Dan Corbiani, and George Bache for helping to build our computing infrastructure. This work was supported by the NNSA Office of Defense Nuclear Nonproliferation Research and Development, U.S. Department of Energy, and Pacific Northwest National Laboratory, which is operated by Battelle Memorial Institute for the U.S. Department of Energy under Contract DE-AC05–76RLO1830. This article has been cleared by PNNL for public release as PNNL-SA-159410.

\fi

\bibliography{references}
\bibliographystyle{acl_natbib}

\onecolumn
\appendix

\section{OSTI Subject Categories}
\label{sec:osti-labels}
\begin{table*}[h!]
    \caption{List of OSTI subject category labels, their description where available, and whether they related directly to the nuclear fuel cycle.}
    \small
    \centering
    \begin{tabular}{rlc|rlc}
        \toprule
        Label & Description               & NFC & Label & Description               & NFC \\
        \midrule
        1     & Coal, Lignite, and Peat   &     & 44    &                           &     \\
        2     & Petroleum                 &     & 45    & Military Technology,      &     \\
        3     & Natural Gas               &     &       & Weaponry, and National    &     \\
        4     & Oil Shales and Tar Sands  &     &       & Defense                   &     \\
        5     & Nuclear Fuels             & Y   & 46    & Instrumentation Related   & Y   \\
        7     & Isotope and Radiation     & Y   &       & To Nuclear Science and    &     \\
              & Sources                   &     &       & Technology                &     \\
        8     & Hydrogen                  &     & 47    & Other Instrumentation     &     \\
        9     & Biomass Fuels             &     & 54    & Environmental Sciences    &     \\
        10    & Synthetic Fuels           &     & 55    &                           &     \\
        11    & Nuclear Fuel Cycle        & Y   & 56    & Biology and Medicine      &     \\
              & and Fuel Materials        &     & 57    &                           &     \\
        12    & Management of Radioactive & Y   & 58    & Geosciences               &     \\
              & and Non-Radioactive       &     & 59    & Basic Biological Sciences &     \\
              & Wastes From Nuclear       &     & 60    & Applied Life Sciences     &     \\
              & Facilities                &     & 61    & Radiation Protection and  &     \\
        13    & Hydro Energy              &     &       & Dosimetry                 &     \\
        14    & Solar Energy              &     & 62    & Radiology and Nuclear     &     \\
        15    & Geothermal Energy         &     &       & Medicine                  &     \\
        16    & Tidal and Wave Power      &     & 63    & Radiation, Thermal, and   &     \\
        17    & Wind Energy               &     &       & Other Environ. Pollutant  &     \\
        20    & Fossil-Fueled Power       &     &       & Effects On Living Orgs.   &     \\
              & Plants                    &     &       & and Biol. Mat.            &     \\
        21    & Specific Nuclear Reactors & Y   & 66    & Physics                   &     \\
              & and Associated Plants     &     & 70    & Plasma Physics and Fusion &     \\
        22    & General Studies of        & Y   &       & Technology                &     \\
              & Nuclear Reactors          &     & 71    & Classical and Quantum     &     \\
        24    & Power Transmission and    &     &       & Mechanics, General        &     \\
              & Distribution              &     &       & Physics                   &     \\
        25    & Energy Storage            &     & 72    & Physics Of Elementary     &     \\
        29    & Enery Planning, Policy,   &     &       & Particles and Fields      &     \\
              & and Economy               &     & 73    & Nuclear Physics and       & Y   \\
        30    & Direct Energy Conversion  &     &       & Radiation Physics         &     \\
        32    & Energy Conservation,      &     & 74    & Atomic and Molecular      &     \\
              & Consumption, and          &     &       & Physics                   &     \\
              & Utilization               &     & 75    & Condensed Matter Physics  &     \\
        33    & Advanced Propulsion       &     &       & Superconductivity and     &     \\
              & Systems                   &     &       & Superfluidity             &     \\
        35    & Arms Control              &     & 77    & Nanoscience and           &     \\
        36    & Material Science          &     &       & Nanotechnology            &     \\
        37    & Inorganic, Organic,       &     & 79    & Astronomy and             &     \\
              & Physical and Analytical   &     &       & Astrophysics              &     \\
              & Chemistry                 &     & 96    & Knowledge Management and  &     \\
        38    & Radiation Chemistry,      & Y   &       & Preservation              &     \\
              & Radiochemistry, and       &     & 97    & Mathematics and Computing &     \\
              & Nuclear Chemistry         &     & 98    & Nuclear Disarmament,      &     \\  
        39    &                           &     &       & Safeguards, and Physical  &     \\
        40    & Chemistry                 &     &       & Protection                &     \\
        42    & Engineering               &     & 99    & General and Miscellaneous &     \\
        43    & Particle Accelerators     &     &       &                           &     \\
        \bottomrule
    \end{tabular}
    \label{tab:osti-labels}
\end{table*}

\newpage
\section{Hyperparameter Tuning}
\label{sec:appendix_hyper}

A hyperparameter tuning experiment is performed on the binary classification task using \textsc{RoBERTa Large}, both with and without domain-adaptive pre-training. We perform a grid search over maximum learning rates of $1\cdot10^{-5}$, $2\cdot10^{-5}$, and $5\cdot10^{-5}$ and minibatch sizes of 16 and 64. Results on the validation set are summarized in Table \ref{tab:hyper_tuning}. Both with and without continued pre-training, a small learning rate and large batch size yield the best loss, though the impact on accuracy and F1 score is both smaller and less clear.

\begin{table*}[h!]
    \caption{Results of a hyperparameter tuning experiment. F1 scores consider NFC-related to be the positive class. The best result for each model is \textbf{bolded}.}
    \small
    \centering
    \begin{tabular}{llllll}
        \toprule
        Model            & Learning Rate   & Batch Size & Accuracy        & F1 score        & Loss            \\
        \midrule
                         & $1\cdot10^{-5}$ & 16         & \textbf{0.9545} & \textbf{0.9537} & 0.1173          \\
                         &                 & 64         & 0.9506          & 0.9502          & \textbf{0.1081} \\
        \textsc{RoBERTa} & $2\cdot10^{-5}$ & 16         & 0.9397          & 0.9409          & 0.1568          \\
        \textsc{Large}   &                 & 64         & 0.9524          & 0.9523          & 0.1118          \\
                         & $5\cdot10^{-5}$ & 16         & 0.9206          & 0.9097          & 0.2260          \\
                         &                 & 64         & 0.9363          & 0.9338          & 0.1699          \\
        \midrule
                         & $1\cdot10^{-5}$ & 16         & \textbf{0.9573} & 0.9568          & 0.1127          \\
                         &                 & 64         & \textbf{0.9573} & \textbf{0.9570} & \textbf{0.0967} \\
        \textsc{RoBERTa} & $2\cdot10^{-5}$ & 16         & 0.9520          & 0.9516          & 0.1340          \\
        \textsc{Large}   &                 & 64         & 0.9557          & 0.9559          & 0.0977          \\
        + OSTI           & $5\cdot10^{-5}$ & 16         & 0.9328          & 0.9279          & 0.2093          \\
                         &                 & 64         & 0.9525          & 0.9518          & 0.1108          \\
        \bottomrule
    \end{tabular}
    \label{tab:hyper_tuning}
\end{table*}


\end{document}